\pdfoutput=1

\documentclass[11pt]{article}

\usepackage{acl}

\usepackage{times}
\usepackage{latexsym}

\usepackage[T1]{fontenc}

\usepackage[utf8]{inputenc}

\usepackage{microtype}
\usepackage{graphicx}
\usepackage{booktabs} 
\usepackage{multirow}
\usepackage{colortbl}
\usepackage{subcaption}
\usepackage[ruled,linesnumbered]{algorithm2e}
\usepackage{algorithmic}
\usepackage{hyperref}
\usepackage{amsmath}
\usepackage{mathtools}
\usepackage{relsize}
\usepackage{float}
\usepackage{microtype}
\SetKwInput{KwInput}{Input}
\SetKwInput{KwOutput}{Output}

\title{Pretrained Transformers Do not Always Improve Robustness}
\title{Pretrained Transformers \textit{do not} Provide more Robust Representation than Traditional Models on Data with Noisy Labels}
\title{Pretrained Transformers Do not Always Improve Robustness}

\author{Swaroop Mishra $\;$ Bhavdeep Singh Sachdeva $\;$ Chitta Baral
\\\\
 Arizona State University }

\begin{document}
\maketitle
\begin{abstract}
Pretrained Transformers (PT) have been shown to improve Out of Distribution (OOD) robustness than traditional models such as Bag of Words (BOW), LSTMs, Convolutional Neural Networks (CNN) powered by Word2Vec and Glove embeddings. How does the robustness comparison hold in a real world setting where some part of the dataset can be noisy? Do PT also provide more robust representation than traditional models on exposure to noisy data? We perform a comparative study on 10 models and find an empirical evidence that PT provide less robust representation than traditional models on exposure to noisy data. We investigate further and augment PT with an adversarial filtering (AF) mechanism that has been shown to improve OOD generalization. However, increase in generalization does not necessarily increase robustness, as we find that noisy data fools the AF method powered by PT.

\end{abstract}

\section{Introduction}
ML models, especially large neural networks, have surpassed humans in a variety of AI benchmarks such as GLUE \cite{wang2018glue}. However these benchmarks all have IID evaluation sets; model performance decreases drastically on testing with OOD data \cite{bras2020adversarial,hendrycks2019benchmarking,  mishra2020dqi, gokhale2022generalized}. 
This raises questions regarding the applicability of ML models \textit{beyond} conventional datasets, demanding the development of better evaluation methods \cite{ribeiro2020beyond, mishra2020our, mishra2021robust} that can reduce the overestimated performance of models and better estimate their capability in real world.


We introduce a robustness metric \textit{\textbf{M}ean \textbf{R}ate of Change of \textbf{A}ccuracy with Change in \textbf{P}oisoining (MRAP)} to measure the real world capability of models by evaluating them on data with varying levels of poisoning. This is an important parameter to evaluate because data poisoning is inevitable in real world due to several potential reasons such as annotation issues due to task hardness/crowdworker error/unclear instruction, lack of resources to verify correctness of data, quick evolution of tasks leaving less time to verify, a translated or domain shifted task where the interpretation flips and there exist some users who deliberately mislead. Also, model performance on poisoned data is a bottleneck to reduce our dependency on the expensive annotation process which also has the risk of generating spurious bias \cite{gururangan2018annotation} . 

Humans can detect changes in their surroundings based on their background knowledge and know to ignore poisoned data. Vision models have not been able to do so as they achieve zero training error on replacing true labels with random labels \cite{zhang2016understanding}. Popular regularization techniques have not been able to control this behavior. 

PT is shown to improve OOD robustness more than any other model \cite{hendrycks2020pretrained}. This is inline with our expectations, as self-supervised pre-training makes the model robust and provides less room to overfit. This paper focuses on studying robustness of a series of models to poisoned data. \textit{Do PT also have higher MRAP than other models?}

We create poisoned data using a simple method of label flipping and experiment across 10 models. 
We find an empirical evidence that PT provide less robust representation than traditional models on exposure to noisy data. Our evidence further suggests that model ranking based on accuracy does not translate to their ranking based on robustness to noisy data. In order to investigate further, we augment PT with an adversarial filtering mechanism AFLite\cite{sakaguchi2020winogrande} that has been shown to improve OOD generalization. Since AFLite was not originally designed to filter poisoned data, we extend it and propose \textit{ Adversarial Filtering of Poisoned Data(AFPLite)}-- a flipped version of AFLite~\cite{sakaguchi2020winogrande}-- that is intended to prune poisoned data. 
We observe that adversarial filtering empowered by PT is not robust against poisoned data; hence increase in generalization does not necessarily increase robustness. 

\section{Experiments}
\label{sec:expt}

\subsection{Models}
We select the same models used in the OOD robustness evaluation work \cite{hendrycks2020pretrained}. The models covers a variety of feature representations.

\textbf{Bag of Words (BOW) :}
It \cite{harris1954distributional} is a primal technique involving neural networks. We choose BOW as it has high bias and low variance

\textbf{Word Embedding:}
Word2Vec \cite{mikolov-etal-2013} converts raw words into vectors using a two-layer neural network. We also include Global Vectors for Word Representation (GloVe) in our study \cite{pennington-etal-2014-glove}.
These embeddings are encoded with word averages \cite{wieting2015universal}, LSTM \cite{Hochreiter-etal-1997} and convolutional neural networks \cite{lecun1995convolutional}.

\textbf{Transformers (PT):}
We use RoBERTA \cite{liu2019roberta} and two versions of the BERT model \cite{devlin2018bert} - BERT-Base and BERT-Large.
\subsection{Datasets}
We experiment over the following sentiment classification datasets.

\textbf{SST-2} \cite{Socher13recursivedeep} is a collection of movie reviews along with human annotations. All the samples are classified as positive or negative.

\textbf{IMDB} \cite{maas-etal-2011-learning} is a collection of highly polarized reviews from the movies database site IMDb. A negative and positive review has scores of $\leq$ 4 and $\geq$ 7 out of 10 respectively. There are no neutral reviews included in dataset.

\subsection{Experimental Setup}
\textbf{Data Poisoning:} We create poisoned data by flipping some labels of the training set, keeping the evaluation set unaltered. We control label flipping at 5 different levels: (i) 0\%, (ii) 30\%, (iii) 50\%, (iv) 70\% and (v) 90\%. We iterate through the dataset, mutating labels at random, till we reach the given threshold. For 70\% and 90\% label mutations, we follow the same procedure in reverse.

\textbf{Implementation:} We leverage AllenNLP \cite{gardner2018allennlp} for implementing models. We lightly tune learning rate and number of epochs to maximize validation performance\footnote{Code is attached in the supplementary material and will be open sourced upon acceptance.}.

\textbf{MRAP:} Let $A_i$ be the accuracies of model $M$ for dataset $D_i$ (the $i^{th}$ variant of dataset $D$) where the percentage of poisoned data is $P_i$, $i \in [1,n]$, $B$ represents the set of available datasets i.e $D \in B$ and $G$ represents the set of available models i.e $M \in G$. We define $MRAP$ and its normalized version $NMRAP$ as:

\begin{equation}
    \resizebox{0.83\linewidth}{0.6cm}{
    \begin{math}
        R_i=
        \begin{cases}
            \frac{P_{i-1}(D_{i-1})-P_i(D_i)}{A_{i-1}(M,D_{i-1})-A_i(M,D_i)} & P_{i-1}(D_{i-1}) < 50\\ 
            \frac{A_i(M,D_i)-A_{i-1}(M,D_{i-1})}{P_{i-1}(D_{i-1})-P_i(D_i)} &  P_{i-1}(D_{i-1}) >=50
        \end{cases}
    \end{math}
}
\end{equation}
\begin{equation}
    MRAP(M,D)=\frac{\sum_{i=2}^n{R_i}}{n-1}
\end{equation}

\begin{equation}
    MRAP(M)=\frac{\sum_{B}MRAP(M,D)}{|B|}
\end{equation}

\begin{equation}
    \resizebox{0.82\linewidth}{0.4cm}{
        \begin{math}
            \begin{aligned}
                \textbf{NMRAP(M)} = \frac{MRAP(M) - \min(MRAP(G))}{ \max(MRAP(G))- \min(MRAP(G))}
            \end{aligned}
        \end{math}
    }
\end{equation}

\section{Results and Analysis}
\label{sec:res}


We calculate NMRAP for all models and explore the following questions while analyzing results.

\begin{figure}[ht]
        \includegraphics[width=\linewidth]{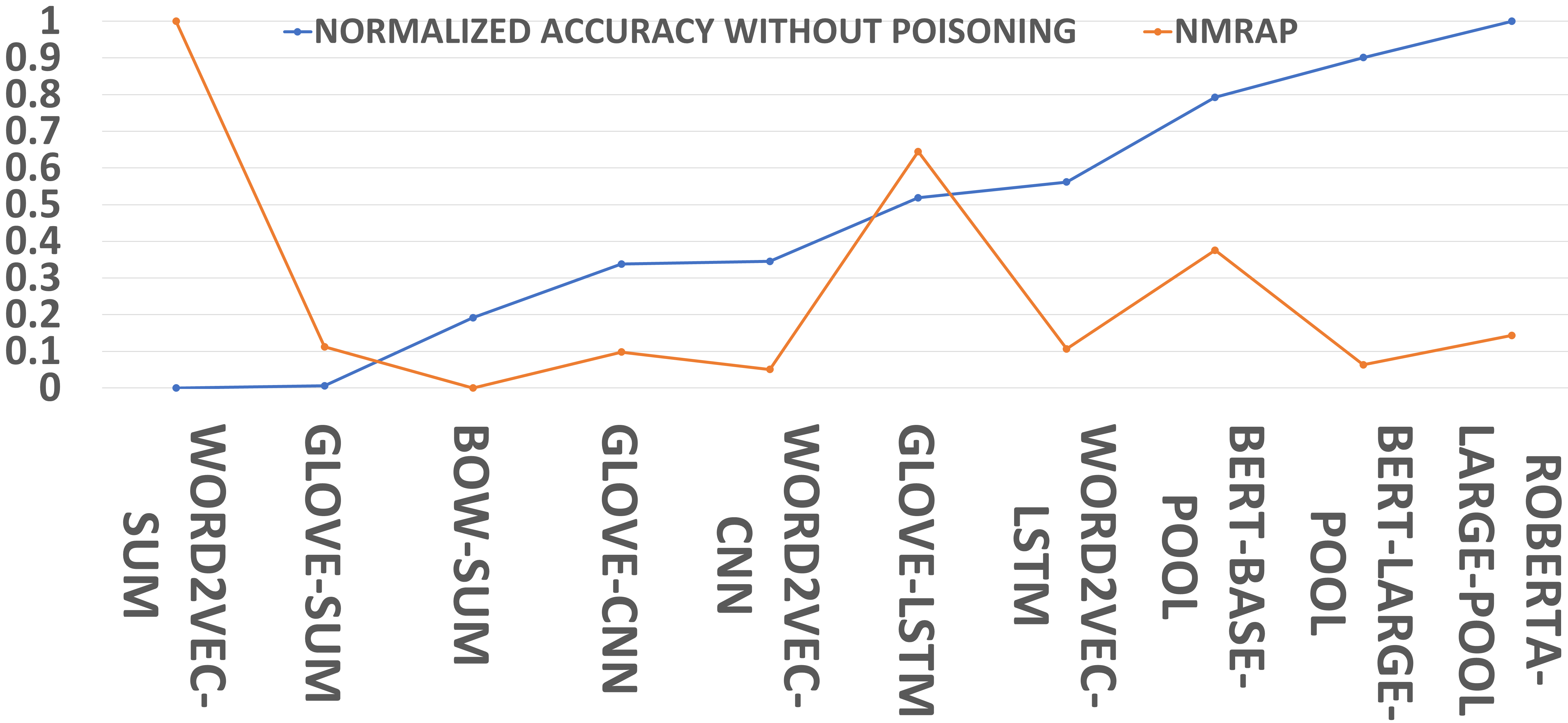} 
        \caption{Ranking of models based on accuracy (normalized to [0,1]) on not poisoned data vs NMRAP.}
        \label{fig:NMRAP}
\end{figure}

\textit{($q_{1}$) Does our advancement in ML models from a simple neural network to Transformers is also reflected in robustness to data poisoning?}
Figure \ref{fig:NMRAP} shows the comparison between normalized accuracy (converted to [0,1] for fair comparison) and NMRAP for 10 models. We arrange models in the increasing order of accuracy, however NMRAP is not monotonically increasing and results in several ranking changes. So, our advancement in ML models (measured using accuracy) is \textbf{not} reflected in robustness to data poisoning. This finding is in-contrast to the emerging new behavior in new models e.g. few shot instruction-following~\cite{mishra2022cross, wei2021finetuned, sanh2021multitask, mishra-etal-2022-reframing, ouyang2022training, parmar2022boxbart}.
We further analyze by segregating models into 4 clusters: Transformers, LSTM, CNN, and others (also called as Naive in various contexts). CNNs and Transformers are seen to perform worse than Naive models and LSTMs (Table \ref{tab:combAvg}). Even though Transformers have higher validation accuracy for each poisoning level, the accuracy perturbs significantly with respect to the poisoning levels; this results in lower robustness which is reflected in NMRAP.

\begin{table}[ht]
    \begin{center}
        \scalebox{0.7} {
            \begin{tabular}{|l|l|l|l|l|}
            \hline
             &  &  &  &  \\
            \multirow{-2}{*}{Pi\textbackslash{}Model} & \multirow{-2}{*}{CNN} & \multirow{-2}{*}{LSTM} & \multirow{-2}{*}{Naive} & \multirow{-2}{*}{Transformers} \\ \hline
             & \cellcolor[HTML]{D0EAD9} & \cellcolor[HTML]{BAE2C7} & \cellcolor[HTML]{E9F5EF} & \cellcolor[HTML]{63BE7B} \\
            \multirow{-2}{*}{0} & \multirow{-2}{*}{\cellcolor[HTML]{D0EAD9}84.29} & \multirow{-2}{*}{\cellcolor[HTML]{BAE2C7}86.25} & \multirow{-2}{*}{\cellcolor[HTML]{E9F5EF}81.96} & \multirow{-2}{*}{\cellcolor[HTML]{63BE7B}94.25} \\
             & \cellcolor[HTML]{FAC0C3} & \cellcolor[HTML]{DFF0E6} & \cellcolor[HTML]{FBE5E7} & \cellcolor[HTML]{FAFCFD} \\
            \multirow{-2}{*}{30} & \multirow{-2}{*}{\cellcolor[HTML]{FAC0C3}67.83} & \multirow{-2}{*}{\cellcolor[HTML]{DFF0E6}82.93} & \multirow{-2}{*}{\cellcolor[HTML]{FBE5E7}75.4} & \multirow{-2}{*}{\cellcolor[HTML]{FAFCFD}80.39} \\
             & \cellcolor[HTML]{F86A6C} & \cellcolor[HTML]{F8696B} & \cellcolor[HTML]{F8696B} & \cellcolor[HTML]{F86D6F} \\
            \multirow{-2}{*}{50} & \multirow{-2}{*}{\cellcolor[HTML]{F86A6C}49.71} & \multirow{-2}{*}{\cellcolor[HTML]{F8696B}49.66} & \multirow{-2}{*}{\cellcolor[HTML]{F8696B}49.49} & \multirow{-2}{*}{\cellcolor[HTML]{F86D6F}50.43} \\
             & \cellcolor[HTML]{FBE3E6} & \cellcolor[HTML]{DEF0E5} & \cellcolor[HTML]{FACACD} & \cellcolor[HTML]{FBECEF} \\
            \multirow{-2}{*}{70} & \multirow{-2}{*}{\cellcolor[HTML]{FBE3E6}75.04} & \multirow{-2}{*}{\cellcolor[HTML]{DEF0E5}83.01} & \multirow{-2}{*}{\cellcolor[HTML]{FACACD}69.83} & \multirow{-2}{*}{\cellcolor[HTML]{FBECEF}77} \\
             & \cellcolor[HTML]{EEF7F3} & \cellcolor[HTML]{BEE3C9} & \cellcolor[HTML]{FBFAFD} & \cellcolor[HTML]{66BF7D} \\
            \multirow{-2}{*}{90} & \multirow{-2}{*}{\cellcolor[HTML]{EEF7F3}81.48} & \multirow{-2}{*}{\cellcolor[HTML]{BEE3C9}85.95} & \multirow{-2}{*}{\cellcolor[HTML]{FBFAFD}79.95} & \multirow{-2}{*}{\cellcolor[HTML]{66BF7D}94.04} \\ \hline
             & \cellcolor[HTML]{D9E1F2} & \cellcolor[HTML]{8EA9DB} & \cellcolor[HTML]{C2D0EB} & \cellcolor[HTML]{CFD9EF} \\
            \multirow{-2}{*}{MRAP} & \multirow{-2}{*}{\cellcolor[HTML]{D9E1F2}87.44} & \multirow{-2}{*}{\cellcolor[HTML]{8EA9DB}245.07} & \multirow{-2}{*}{\cellcolor[HTML]{C2D0EB}136.12} & \multirow{-2}{*}{\cellcolor[HTML]{CFD9EF}110.02} \\ \hline
             & \cellcolor[HTML]{D9E1F2} & \cellcolor[HTML]{8EA9DB} & \cellcolor[HTML]{C2D0EB} & \cellcolor[HTML]{CFDAEF} \\
            \multirow{-2}{*}{NMRAP} & \multirow{-2}{*}{\cellcolor[HTML]{D9E1F2}0} & \multirow{-2}{*}{\cellcolor[HTML]{8EA9DB}1} & \multirow{-2}{*}{\cellcolor[HTML]{C2D0EB}0.31} & \multirow{-2}{*}{\cellcolor[HTML]{CFDAEF}0.14} \\ \hline
            \end{tabular}
        }
    \caption{Category-wise validation accuracy of models across varying degrees of label mutation.}
    \label{tab:combAvg}
    \end{center}
    \vspace{-7mm}
\end{table}

\textit{($q_{2}$) Does pre-training with a greater quantity and diversity of data improve robustness to data poisoning?}
Table \ref{tab:combAvg} --where Pre-trained Transformers are outperformed by simpler models-- indicates that pre-training does \textbf{not necessarily} improve robustness to data poisoning. Within Transformers, BERT-BASE-POOL outperforms ROBERTA-LARGE-POOL (Figure \ref{fig:NMRAP}). RoBERTA is pretrained with a greater quantity and diversity of data than BERT. This signals that pre-training with a greater quantity and diversity of data does \textbf{not necessarily} improve this aspect of robustness.

\begin{figure}[ht]
        \includegraphics[width=\linewidth]{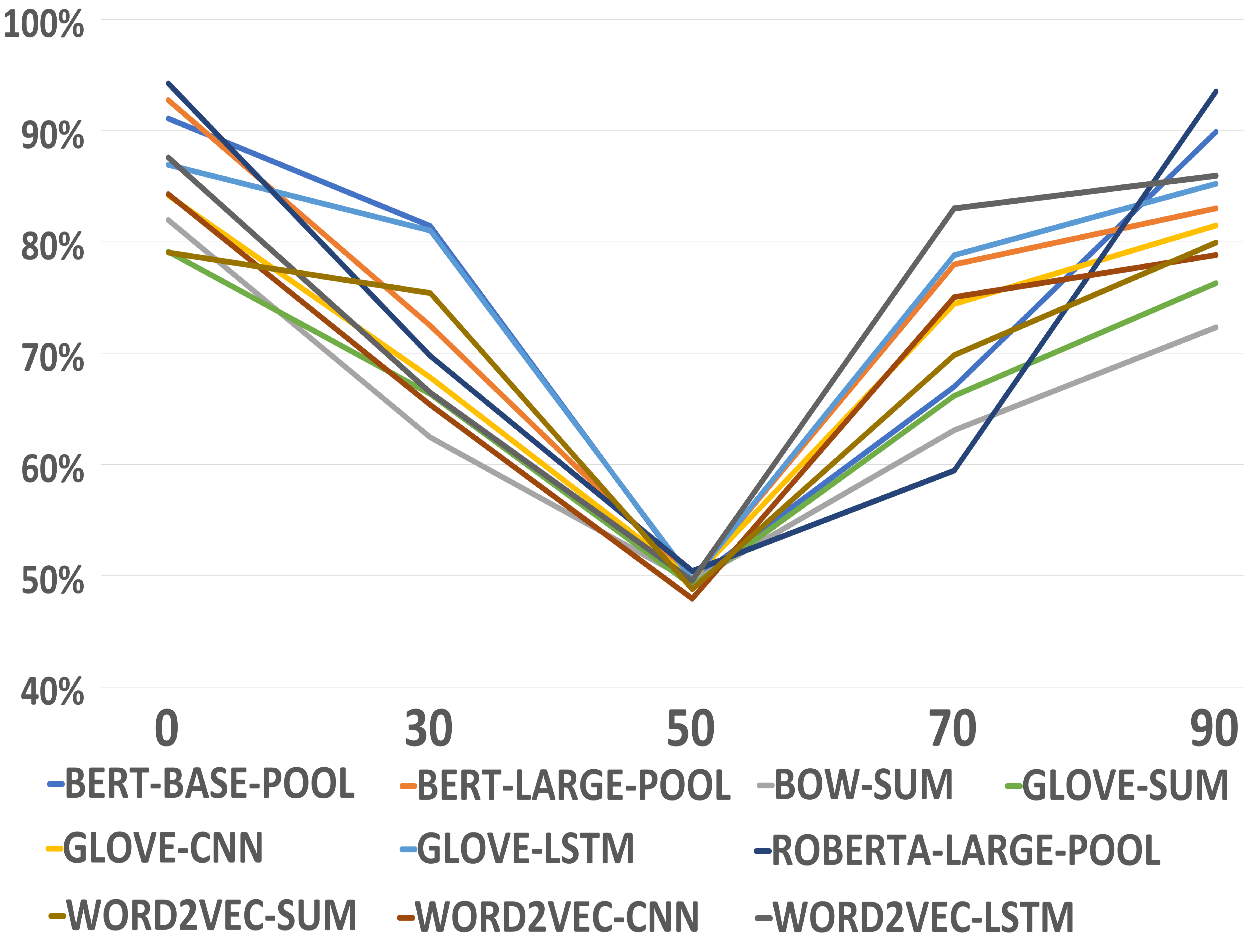} 
        \vspace{-4mm}
        \caption{Validation accuracy (Y axis) vs percentage of label mutation (X axis) for 10 models}
        \label{fig:diff}
\end{figure}
\begin{figure}[ht]
\centering
  \includegraphics[width=\linewidth]{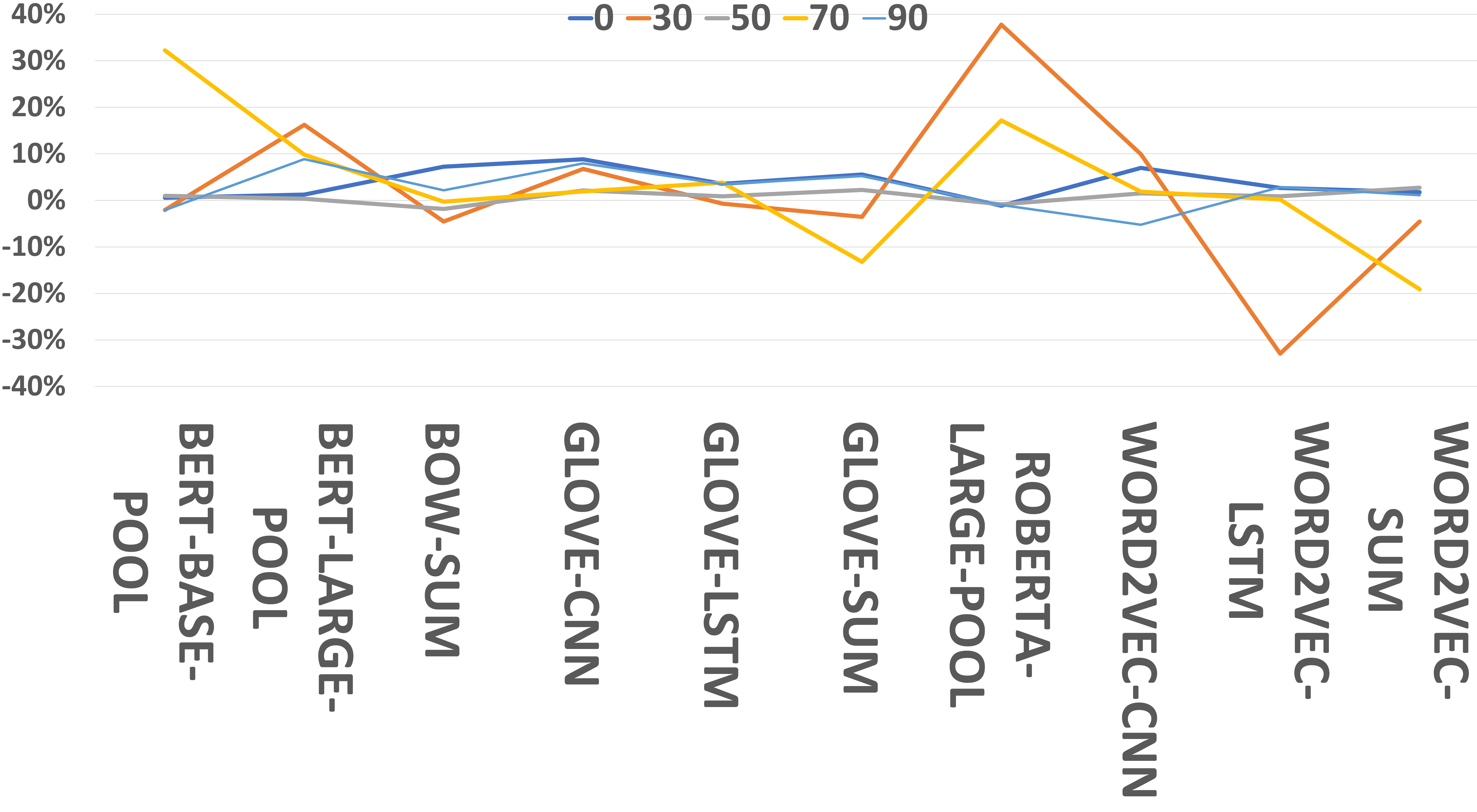}
  \caption{Difference in validation performance of IMDB and SST-2 datasets for 10 models}
  \label{fig:var}
\end{figure}

\textit{($q_{3}$) How are the models sensitive to the percentage of poisoning within and across datasets?}
Figure \ref{fig:diff} shows that, model accuracy initially decreases as we increase the percentage of label mutations. However, this trend reverses beyond 50\% mutation- the accuracy increases- because the label interpretation is flipped when over 50\% of the labels are flipped. WORD2VEC-SUM shows a minimal change in the validation accuracy at all poisoning levels, thus has the highest NMRAP score. 
Figure \ref{fig:var} further illustrates the sensitivity of models with respect to datasets; Transformers have relatively higher variation than other architectures. Currently MRAP and NMRAP does not take variance across datasets into consideration, however those can be integrated in a future metric for fine grained analysis of robustness.

\begin{figure}[ht]
\centering
  \includegraphics[width=\columnwidth]{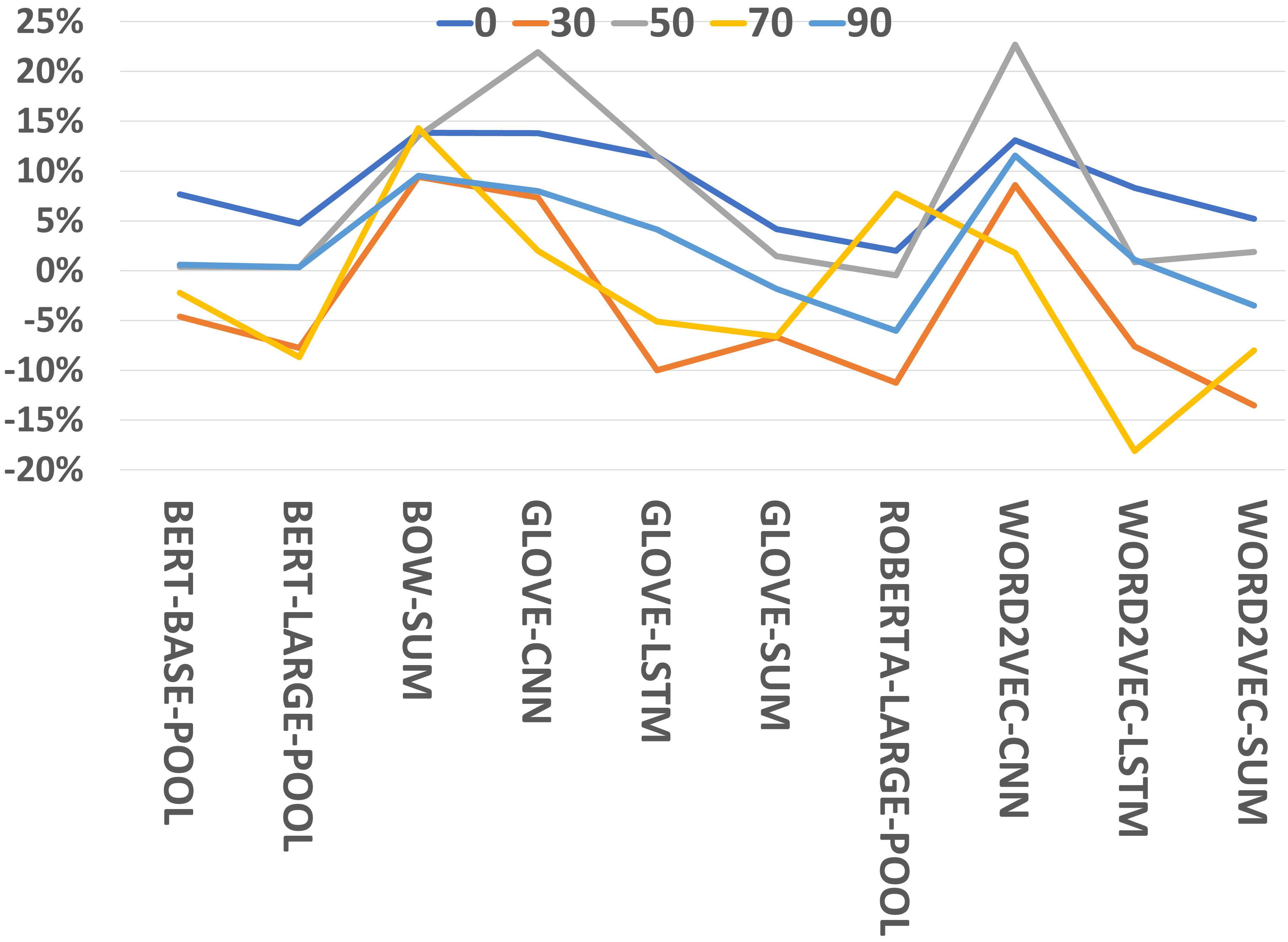}
  \caption{Difference in training and validation accuracy (Y axis) of 10 models (X axis) for varying percentage of label mutation (represented in various colors)}
  \label{fig:vara}
\end{figure}

\textit{($q_{4}$) How does the generalization gap varies across models?}
The difference between training and testing accuracy represents the generalization gap. Higher gap indicates overfitting and so undesirable. Figure \ref{fig:vara} shows that CNNs and Transformers have highest and lowest generalization gap respectively. This characteristics can also be incorporated in a future metric, specially for applications where getting a poison free development set is less likely and so the primary indicator of model capability is training set performance.
\vspace{-2mm}
\section{Adversarial Filtering of Noisy Labels}
\subsection{Algorithm}
We design AFPLite (Algorithm \ref{algo:one}) based on the results of Section \ref{sec:res} that shows how noisy labels pose challenges to PT. Here, we use RoBERTA as PT. The key intuition behind AFPLite is that samples with flipped labels will be hard to solve using linear models on top of RoBERTA embeddings. This is exact opposite of AFLite's intuition, i.e., samples containing spurious bias -- that provides shortcuts to models for solving-- will be easy to solve using linear models on top of RoBERTA embeddings.

\paragraph{Formalization:} Let $D$ represents dataset, $s$ represent samples, $M$ be the set of models (SVM and Logistic Regression) , $S$ be the pruned set, $E(s)$, $C(s)$ and $P(s)$ be the evaluation score, correct prediction score and predictability score of each sample $s$ respectively.

\begin{algorithm}[ht]
    \small
        \KwInput{Dataset $D$, \textbf{Hyper-Parameters}: $m$, $n$, $t$, $k$ and $tau$}
        \KwOutput{Pruned dataset $S$}
        Fine tune RoBERTA on $10$ \% of $D$ and get embeddings for rest of $D$\; 
        $S=D$ - 10 \% of $D$ used in fine tuning\;
        \While{|S|> n} {
             \ForAll{$i \in m$}{
                 Randomly select train set of size $t$ from $S$ \;
                 Train $M$ on $t$ and test on rest of $S$ i.e. $V$ \;
                 \ForAll{$j \in M$}{
                     \ForAll{$s \in V$}{ $E(s) = E(s)+1$\; \If {model prediction is correct}
                     {$C(s) = C(s)+1$}}
                 }
             }
              \ForAll{$s \in S$}{$P(s)=C(s)/E(s)$}
              Sort $S$ descendingly based on $P(s)$ and delete upto $k$ instances from $S$ for which $P(s)<tau$\;
         }
     \caption{AFPLite}
     \label{algo:one}
\end{algorithm}

\subsection{Results and Analysis}
\begin{figure}[ht]
        \includegraphics[width=\linewidth]{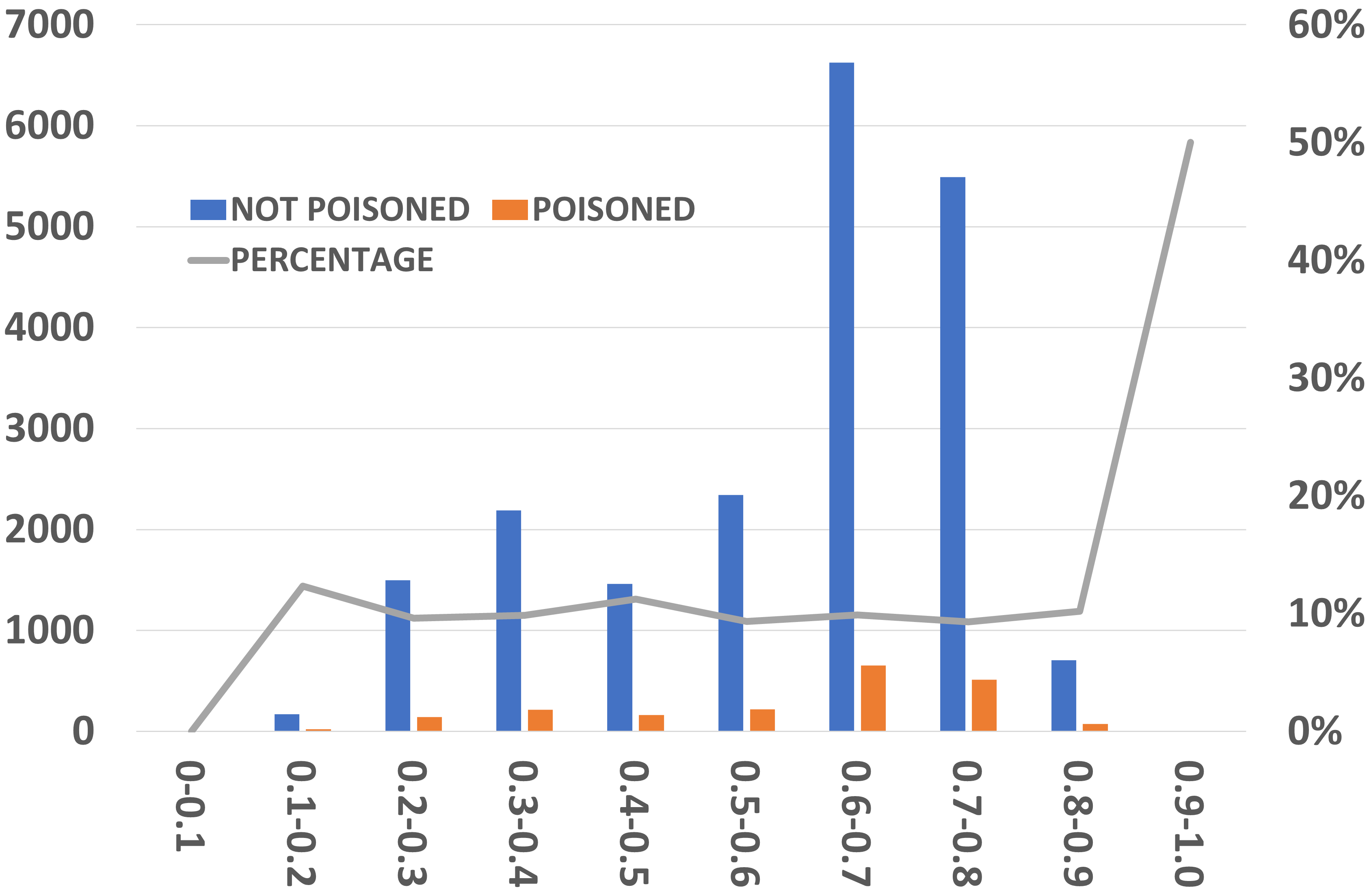} 
        \caption{Ratio of the Poisoned data and not poisoned data (gray line) for each bin of Predictability score (P(s)) on 10\% poisoned data. Bar chart represents total data in each bin. Ideally the ratio should have higher value for bins closer to the origin, however here the flat line shows that AFPLite is fooled with poisoned data.}
        \label{fig:10_Percent_Data}
\end{figure}

A key hyperparameter in Algorithm \ref{algo:one} is $tau$ which defines the upper threshold for filtering. Since samples with noisy labels are hard samples, we start with $tau$ as 0.1 and keep on increasing it til 1, however we could never filter only poisoned data. Filtered data always contain poisoned and not poisoned data uniformly (Figure \ref{fig:10_Percent_Data}). We also observe the same distribution for varying percentages of poisoning (Table \ref{tab:Raio}). Since Adversarial filtering (empowered by PT i.e. RoBERTA embeddings) is unable to distinguish poisoned and nonpoisoned data, data poisoning is indeed a challenge for PT.

\begin{table}[ht]
\centering
\scriptsize
\begin{tabular}{|l|llll|}
\hline
P(s)\textbackslash{}$P_i$ & 1 & 5 & 10 & 50  \\ \hline
0-0.1   & \cellcolor[HTML]{FFF2CC}0.00 & \cellcolor[HTML]{FFF2CC}0.00 & \cellcolor[HTML]{FFF2CC}0.00  & \cellcolor[HTML]{FFF2CC}0.00   \\
0.1-0.2 & \cellcolor[HTML]{FFF1C6}3.25 & \cellcolor[HTML]{8EA9DB}5.32 & \cellcolor[HTML]{8EA9DB}12.35 & \cellcolor[HTML]{B1C3E6}100.00 \\
0.2-0.3 & \cellcolor[HTML]{B4C6E7}0.90 & \cellcolor[HTML]{C0CEEB}4.40 & \cellcolor[HTML]{D2DCF0}9.62  & \cellcolor[HTML]{C6D3ED}88.24  \\
0.3-0.4 & \cellcolor[HTML]{BDCCEA}0.84 & \cellcolor[HTML]{D2DCF0}4.07 & \cellcolor[HTML]{CCD7EE}9.87  & \cellcolor[HTML]{CCD7EE}85.20  \\
0.4-0.5 & \cellcolor[HTML]{D9E1F2}0.63 & \cellcolor[HTML]{D3DCF0}4.05 & \cellcolor[HTML]{AABEE4}11.23 & \cellcolor[HTML]{D3DDF1}81.32  \\
0.5-0.6 & \cellcolor[HTML]{8EA9DB}1.17 & \cellcolor[HTML]{B3C5E7}4.64 & \cellcolor[HTML]{D8E1F2}9.35  & \cellcolor[HTML]{D3DDF1}81.20  \\
0.6-0.7 & \cellcolor[HTML]{A5BBE2}1.01 & \cellcolor[HTML]{B0C2E6}4.70 & \cellcolor[HTML]{CBD7EE}9.88  & \cellcolor[HTML]{D9E1F2}77.73  \\
0.7-0.8 & \cellcolor[HTML]{C2D0EB}0.80 & \cellcolor[HTML]{A8BCE3}4.85 & \cellcolor[HTML]{D9E1F2}9.31  & \cellcolor[HTML]{8EA9DB}119.05 \\
0.8-0.9 & \cellcolor[HTML]{BCCCEA}0.84 & \cellcolor[HTML]{97B0DE}5.17 & \cellcolor[HTML]{C3D1EC}10.21 & \cellcolor[HTML]{FFF2CC}0.00   \\
0.9-1.0 & \cellcolor[HTML]{FFF2CC}0.00 & \cellcolor[HTML]{D9E1F2}3.92 & \cellcolor[HTML]{FFD966}50.00 & \cellcolor[HTML]{FFF2CC}0.00   \\ \hline
\end{tabular}
\caption{ Ratio of poisoned data to not poisoned data in AFPLite, Columns represent percentages of poisoning and rows represent predictability scores.}
\label{tab:Raio}
\end{table}

\section{Conclusion}
Considering the model performance inflation caused by accuracy, we propose MRAP to measure robustness of models in handling poisoned data, a key skill required in the real world for various reasons. We experiment across 10 models and find an evidence where LSTMs and other models outperform PT in robustness to noisy data. In order to understand the failure mode of PT, we extend AFLite and build AFPLite - a flipped version of AFLite- that is meant to prune poisoned data. We show that adversarial filtering (empowered by PT) which is known to improve OOD generalization does not necessarily improve robustness as a simple data poisoning method fools adversarial filtering. We analyze several concepts related to robustness such as generalization, pre-training and sensitivity. A potential future extension is to analyze our evidence further by performing large scale experiment on more datasets and models. We hope our insights from model comparison will help the community develop better real world models/ensembles. 
\section{Limitations}
We experiment with only 2 datasets. However, note that the goal of this paper is to justify a negative result. As in logic and mathematics, only one instance is enough to prove that something is not True (in contrast to the exhaustive evaluation required to proof something is True).

Noisy data in real world can be more complex than label flipping. In this paper, all our experiments are limited to label flipping. However, the goal of this paper is to show a negative finding. We feel its actually a strength of the paper to show the negative finding with such a simple technique.

\bibliography{anthology,custom}
\bibliographystyle{acl_natbib}

\appendix
\section{Supplemental Material}
This supplementary section is provided to shed some more light on the experiments presented in the paper.
\subsubsection{Models Accuracy}
The table \ref{tab:overallResults} provides more detail for the model's accuracy with respect to the amount of poisoning in the data as discussed in $q_{3}$ of Results and Analysis Section. 
\begin{figure}
    \centering
    \begin{subfigure}[ht]{.5\textwidth}
        \includegraphics[width=0.9\linewidth,height=3.2cm]{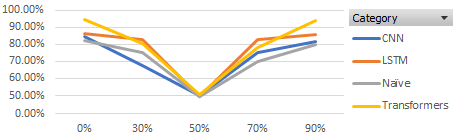}
        \caption{Validation accuracy of each model category vs. level of label mutations}
        \label{fig:comb}
    \end{subfigure}
\end{figure}
    
\begin{table*}[ht]
\centering
\resizebox{\textwidth}{!}{
\begin{tabular}{|l|l|l|l|l|l|l|l|l|l|l|}
\hline
~  & bert-base-pool                              & bert-large-pool                             & bow-sum                                     & glove-cnn                                   & glove-lstm                                  & glove-sum                                   & roberta-large-pool                          & word2vec-cnn                                & word2vec-lstm                               & word2vec-sum                                 \\
\hline
0  & {\cellcolor[rgb]{0.537,0.808,0.612}}91.36\% & {\cellcolor[rgb]{0.439,0.769,0.529}}93.65\% & {\cellcolor[rgb]{0.784,0.906,0.824}}85.58\% & {\cellcolor[rgb]{0.565,0.816,0.631}}90.79\% & {\cellcolor[rgb]{0.773,0.902,0.812}}85.91\% & {\cellcolor[rgb]{0.863,0.937,0.89}}83.75\%  & {\cellcolor[rgb]{0.388,0.745,0.482}}94.85\% & {\cellcolor[rgb]{0.616,0.839,0.678}}89.52\% & {\cellcolor[rgb]{0.757,0.894,0.8}}86.25\%   & {\cellcolor[rgb]{0.988,0.988,1}}80.76\%      \\
30 & {\cellcolor[rgb]{0.984,0.98,0.992}}80.39\%  & {\cellcolor[rgb]{0.682,0.867,0.737}}87.97\% & {\cellcolor[rgb]{0.976,0.694,0.706}}64.72\% & {\cellcolor[rgb]{0.984,0.843,0.855}}72.92\% & {\cellcolor[rgb]{0.953,0.973,0.969}}81.67\% & {\cellcolor[rgb]{0.984,0.961,0.973}}79.27\% & {\cellcolor[rgb]{0.655,0.855,0.714}}88.65\% & {\cellcolor[rgb]{0.98,0.839,0.851}}72.78\%  & {\cellcolor[rgb]{0.898,0.953,0.922}}82.93\% & {\cellcolor[rgb]{0.984,0.98,0.992}}80.53\%   \\
50 & {\cellcolor[rgb]{0.973,0.427,0.435}}50.00\% & {\cellcolor[rgb]{0.973,0.443,0.451}}50.86\% & {\cellcolor[rgb]{0.973,0.435,0.443}}50.40\% & {\cellcolor[rgb]{0.973,0.455,0.463}}51.55\% & {\cellcolor[rgb]{0.973,0.435,0.443}}50.52\% & {\cellcolor[rgb]{0.973,0.443,0.451}}50.96\% & {\cellcolor[rgb]{0.973,0.443,0.451}}50.86\% & {\cellcolor[rgb]{0.973,0.412,0.42}}49.08\%  & {\cellcolor[rgb]{0.973,0.412,0.42}}49.14\%  & {\cellcolor[rgb]{0.973,0.447,0.455}}51.20\%  \\
70 & {\cellcolor[rgb]{0.89,0.949,0.914}}83.12\%  & {\cellcolor[rgb]{0.624,0.843,0.686}}89.35\% & {\cellcolor[rgb]{0.976,0.667,0.678}}63.23\% & {\cellcolor[rgb]{0.984,0.898,0.91}}75.89\%  & {\cellcolor[rgb]{0.965,0.98,0.98}}81.35\%   & {\cellcolor[rgb]{0.984,0.925,0.937}}77.43\% & {\cellcolor[rgb]{0.973,0.443,0.451}}50.86\% & {\cellcolor[rgb]{0.984,0.906,0.918}}76.45\% & {\cellcolor[rgb]{0.89,0.949,0.918}}83.09\%  & {\cellcolor[rgb]{0.984,0.973,0.984}}80.07\%  \\
90 & {\cellcolor[rgb]{0.647,0.851,0.706}}88.85\% & {\cellcolor[rgb]{0.494,0.788,0.573}}92.44\% & {\cellcolor[rgb]{0.984,0.851,0.863}}73.43\% & {\cellcolor[rgb]{0.706,0.875,0.757}}87.42\% & {\cellcolor[rgb]{0.702,0.875,0.753}}87.55\% & {\cellcolor[rgb]{0.984,0.984,0.996}}80.64\% & {\cellcolor[rgb]{0.424,0.761,0.514}}94.04\% & {\cellcolor[rgb]{0.988,0.988,1}}80.76\%     & {\cellcolor[rgb]{0.71,0.875,0.761}}87.37\%  & {\cellcolor[rgb]{0.976,0.984,0.988}}81.10\% \\
\hline
\end{tabular}
}
\caption{Maximum validation accuracy of various models with different levels of label corruption}
\label{tab:overallResults}
\end{table*}

\subsubsection{Model Categorization}
Table \ref{tab:combined21} depicts how different models are combined and averaged to produce the results in Results and Analysis section. Table \ref{tab:generalisationGap} further illustrates the generalization gap between training and dev set as discussed in Results and Analysis section.
\begin{table*}[h]
\begin{center}
\begin{tabular}{|l|l|l|l|l|l|l|l|}
\hline
              &         &              & \multicolumn{5}{l|}{Corruption Percentage
  Percentage}                                                                                                                                                                               \\
Embedder      & Seq2Vec & Category     & 0                                           & 30                                          & 50                                          & 70                                          & 90                 \\
\hline
bow           & sum     & Naïve        & {\cellcolor[rgb]{0.792,0.91,0.831}}81.96\%  & {\cellcolor[rgb]{0.98,0.71,0.718}}62.43\%   & {\cellcolor[rgb]{0.973,0.443,0.451}}49.49\% & {\cellcolor[rgb]{0.98,0.722,0.733}}63.08\%  & {\cellcolor[rgb]{0.984,0.914,0.925}}72.34\%  \\
word2vec      & sum     & Naïve        & {\cellcolor[rgb]{0.886,0.949,0.914}}79.04\% & {\cellcolor[rgb]{0.984,0.976,0.988}}75.40\% & {\cellcolor[rgb]{0.973,0.427,0.435}}48.80\% & {\cellcolor[rgb]{0.984,0.863,0.875}}69.83\% & {\cellcolor[rgb]{0.855,0.937,0.886}}79.95\%  \\
glove         & sum     & Naïve        & {\cellcolor[rgb]{0.882,0.945,0.91}}79.14\%  & {\cellcolor[rgb]{0.98,0.788,0.8}}66.33\%    & {\cellcolor[rgb]{0.973,0.435,0.443}}49.12\% & {\cellcolor[rgb]{0.98,0.788,0.796}}66.18\%  & {\cellcolor[rgb]{0.976,0.984,0.988}}76.31\%  \\
word2vec      & lstm    & LSTM         & {\cellcolor[rgb]{0.651,0.851,0.71}}86.25\%  & {\cellcolor[rgb]{0.761,0.898,0.804}}82.93\% & {\cellcolor[rgb]{0.973,0.435,0.443}}49.14\% & {\cellcolor[rgb]{0.757,0.894,0.8}}83.01\%   & {\cellcolor[rgb]{0.663,0.855,0.718}}85.95\%  \\
glove         & lstm    & LSTM         & {\cellcolor[rgb]{0.667,0.859,0.725}}85.74\% & {\cellcolor[rgb]{0.812,0.918,0.851}}81.27\% & {\cellcolor[rgb]{0.973,0.447,0.455}}49.66\% & {\cellcolor[rgb]{0.894,0.953,0.918}}78.81\% & {\cellcolor[rgb]{0.682,0.867,0.737}}85.23\%  \\
word2vec      & cnn     & CNN          & {\cellcolor[rgb]{0.714,0.878,0.765}}84.29\% & {\cellcolor[rgb]{0.98,0.769,0.78}}65.36\%   & {\cellcolor[rgb]{0.973,0.412,0.42}}47.95\%  & {\cellcolor[rgb]{0.984,0.969,0.98}}75.04\%  & {\cellcolor[rgb]{0.894,0.949,0.918}}78.83\%  \\
glove         & cnn     & CNN          & {\cellcolor[rgb]{0.718,0.878,0.769}}84.18\% & {\cellcolor[rgb]{0.98,0.82,0.831}}67.83\%   & {\cellcolor[rgb]{0.973,0.447,0.455}}49.71\% & {\cellcolor[rgb]{0.984,0.957,0.969}}74.44\% & {\cellcolor[rgb]{0.808,0.918,0.843}}81.48\%  \\
bert-base     & pool    & Transformers & {\cellcolor[rgb]{0.486,0.784,0.565}}91.36\% & {\cellcolor[rgb]{0.843,0.929,0.875}}80.39\% & {\cellcolor[rgb]{0.973,0.443,0.451}}49.51\% & {\cellcolor[rgb]{0.98,0.804,0.812}}66.99\%  & {\cellcolor[rgb]{0.565,0.82,0.635}}88.85\%   \\
bert-large    & pool    & Transformers & {\cellcolor[rgb]{0.439,0.769,0.525}}92.74\% & {\cellcolor[rgb]{0.984,0.918,0.929}}72.47\% & {\cellcolor[rgb]{0.973,0.447,0.455}}49.73\% & {\cellcolor[rgb]{0.922,0.961,0.941}}77.98\% & {\cellcolor[rgb]{0.757,0.894,0.8}}83.02\%    \\
roberta-large & pool    & Transformers & {\cellcolor[rgb]{0.388,0.745,0.482}}94.25\% & {\cellcolor[rgb]{0.984,0.859,0.871}}69.75\% & {\cellcolor[rgb]{0.973,0.463,0.471}}50.43\% & {\cellcolor[rgb]{0.973,0.471,0.478}}50.86\% & {\cellcolor[rgb]{0.396,0.749,0.49}}94.04\% \\ 
\hline
\end{tabular}
\end{center}
\caption{Combination of given models into broader categories i.e. Transformers, Convolutional neural network(CNN), Long short-term memory(LSTM) and others labeled as naive.}
\label{tab:combined21}
\end{table*}



\subsection{Infrastructure Used}
All the experiments were conducted on "TeslaV100-SXM2-16GB"; CPU cores per node 20; CPU memory per node: 95,142 MB; CPU memory per core: 4,757 MB. This configuration is not a necessity for these experiments as we ran our operations with NVIDIA Quadro RTX 4000 as well with lesser memory.

\begin{table*}[ht]
\centering
\scalebox{0.7}{
\begin{tabular}{|l|l|l|l|l|l|l|}
\hline
& & \multicolumn{1}{r|}{0} & \multicolumn{1}{r|}{30} & \multicolumn{1}{r|}{50} & \multicolumn{1}{r|}{70} & \multicolumn{1}{r|}{90} \\
\hline
\multirow{10}{*}{IMDB}  & bert-base-pool     & {\cellcolor[rgb]{0.918,0.961,0.937}}7.91\%  & {\cellcolor[rgb]{0.984,0.922,0.933}}-0.74\%  & {\cellcolor[rgb]{0.984,0.937,0.949}}-0.19\% & {\cellcolor[rgb]{0.98,0.827,0.839}}-4.86\%   & {\cellcolor[rgb]{0.98,0.988,0.992}}2.72\%    \\
                        & bert-large-pool    & {\cellcolor[rgb]{0.988,0.988,1}}1.99\%      & {\cellcolor[rgb]{0.98,0.765,0.773}}-7.58\%   & {\cellcolor[rgb]{0.984,0.941,0.953}}0.02\%  & {\cellcolor[rgb]{0.98,0.776,0.788}}-7.05\%   & {\cellcolor[rgb]{0.973,0.984,0.988}}3.43\%   \\
                        & bow-sum            & {\cellcolor[rgb]{0.847,0.933,0.878}}13.69\% & {\cellcolor[rgb]{0.761,0.898,0.804}}20.44\%  & {\cellcolor[rgb]{0.749,0.894,0.796}}21.44\% & {\cellcolor[rgb]{0.651,0.855,0.71}}29.26\%   & {\cellcolor[rgb]{0.878,0.945,0.906}}11.03\%  \\
                        & glove-cnn          & {\cellcolor[rgb]{0.902,0.953,0.925}}9.21\%  & {\cellcolor[rgb]{0.733,0.886,0.78}}22.62\%   & {\cellcolor[rgb]{0.42,0.757,0.51}}48.22\%   & {\cellcolor[rgb]{0.722,0.882,0.769}}23.68\%  & {\cellcolor[rgb]{0.878,0.945,0.906}}10.93\%  \\
                        & glove-lstm         & {\cellcolor[rgb]{0.91,0.957,0.933}}8.40\%   & {\cellcolor[rgb]{0.98,0.745,0.757}}-8.32\%   & {\cellcolor[rgb]{0.682,0.867,0.737}}26.94\% & {\cellcolor[rgb]{0.98,0.769,0.776}}-7.42\%   & {\cellcolor[rgb]{0.984,0.98,0.992}}1.71\%    \\
                        & glove-sum          & {\cellcolor[rgb]{0.984,0.914,0.925}}-1.19\% & {\cellcolor[rgb]{0.98,0.788,0.8}}-6.56\%     & {\cellcolor[rgb]{0.984,0.937,0.949}}-0.17\% & {\cellcolor[rgb]{0.984,0.98,0.992}}1.70\%    & {\cellcolor[rgb]{0.98,0.729,0.741}}-9.04\%   \\
                        & roberta-large-pool & {\cellcolor[rgb]{0.988,0.988,1}}1.99\%      & {\cellcolor[rgb]{0.973,0.412,0.42}}-22.83\%  & {\cellcolor[rgb]{0.984,0.941,0.949}}-0.04\% & {\cellcolor[rgb]{0.831,0.925,0.863}}14.92\%  & {\cellcolor[rgb]{0.984,0.843,0.855}}-4.21\%  \\
                        & word2vec-cnn       & {\cellcolor[rgb]{0.906,0.957,0.929}}8.75\%  & {\cellcolor[rgb]{0.827,0.925,0.859}}15.24\%  & {\cellcolor[rgb]{0.388,0.745,0.482}}50.46\% & {\cellcolor[rgb]{0.745,0.89,0.788}}21.91\%   & {\cellcolor[rgb]{0.71,0.878,0.761}}24.56\%   \\
                        & word2vec-lstm      & {\cellcolor[rgb]{0.937,0.969,0.957}}6.23\%  & {\cellcolor[rgb]{0.984,0.949,0.961}}0.40\%   & {\cellcolor[rgb]{0.984,0.945,0.957}}0.23\%  & {\cellcolor[rgb]{0.973,0.459,0.467}}-20.67\% & {\cellcolor[rgb]{0.984,0.918,0.929}}-0.93\%  \\
                        & word2vec-sum       & {\cellcolor[rgb]{0.973,0.984,0.984}}3.48\%  & {\cellcolor[rgb]{0.976,0.655,0.663}}-12.29\% & {\cellcolor[rgb]{0.984,0.933,0.945}}-0.34\% & {\cellcolor[rgb]{0.953,0.976,0.969}}4.95\%   & {\cellcolor[rgb]{0.98,0.804,0.812}}-5.94\%   \\
                        \hline
  ~                       &                    & ~                                           & ~                                            & ~                                           & ~                                            & ~
                        \\
                        \hline
\multirow{10}{*}{SST-2} & bert-base-pool     & {\cellcolor[rgb]{0.757,0.894,0.8}}7.40\%    & {\cellcolor[rgb]{0.976,0.69,0.702}}-8.49\%   & {\cellcolor[rgb]{0.976,0.984,0.988}}0.94\%  & {\cellcolor[rgb]{0.984,0.984,0.996}}0.44\%   & {\cellcolor[rgb]{0.984,0.918,0.929}}-1.55\%  \\
                        & bert-large-pool    & {\cellcolor[rgb]{0.831,0.925,0.867}}5.13\%  & {\cellcolor[rgb]{0.973,0.459,0.467}}-15.60\% & {\cellcolor[rgb]{0.984,0.949,0.961}}-0.66\% & {\cellcolor[rgb]{0.973,0.514,0.522}}-13.95\% & {\cellcolor[rgb]{0.984,0.976,0.988}}0.26\%   \\
                        & bow-sum            & {\cellcolor[rgb]{0.533,0.804,0.608}}14.01\% & {\cellcolor[rgb]{0.984,0.918,0.929}}-1.64\%  & {\cellcolor[rgb]{0.816,0.922,0.851}}5.61\%  & {\cellcolor[rgb]{0.984,0.949,0.961}}-0.64\%  & {\cellcolor[rgb]{0.737,0.886,0.784}}7.99\%   \\
                        & glove-cnn          & {\cellcolor[rgb]{0.518,0.8,0.596}}14.41\%   & {\cellcolor[rgb]{0.965,0.98,0.976}}1.30\%    & {\cellcolor[rgb]{0.388,0.745,0.482}}18.16\% & {\cellcolor[rgb]{0.98,0.784,0.796}}-5.63\%   & {\cellcolor[rgb]{0.749,0.89,0.792}}7.64\%    \\
                        & glove-lstm         & {\cellcolor[rgb]{0.573,0.82,0.639}}12.84\%  & {\cellcolor[rgb]{0.976,0.604,0.612}}-11.15\% & {\cellcolor[rgb]{0.882,0.945,0.91}}3.70\%   & {\cellcolor[rgb]{0.98,0.776,0.784}}-5.92\%   & {\cellcolor[rgb]{0.831,0.925,0.867}}5.12\%   \\
                        & glove-sum          & {\cellcolor[rgb]{0.847,0.933,0.878}}4.74\%  & {\cellcolor[rgb]{0.973,0.451,0.463}}-15.78\% & {\cellcolor[rgb]{0.965,0.98,0.98}}1.25\%    & {\cellcolor[rgb]{0.973,0.502,0.51}}-14.33\%  & {\cellcolor[rgb]{0.984,0.871,0.882}}-3.04\%  \\
                        & roberta-large-pool & {\cellcolor[rgb]{0.937,0.969,0.957}}2.01\%  & {\cellcolor[rgb]{0.984,0.98,0.992}}0.34\%    & {\cellcolor[rgb]{0.984,0.941,0.949}}-0.93\% & {\cellcolor[rgb]{0.988,0.988,1}}0.55\%       & {\cellcolor[rgb]{0.98,0.714,0.722}}-7.87\%   \\
                        & word2vec-cnn       & {\cellcolor[rgb]{0.557,0.816,0.627}}13.28\% & {\cellcolor[rgb]{0.804,0.914,0.843}}5.93\%   & {\cellcolor[rgb]{0.533,0.804,0.608}}13.90\% & {\cellcolor[rgb]{0.98,0.82,0.831}}-4.56\%    & {\cellcolor[rgb]{0.761,0.898,0.804}}7.30\%   \\
                        & word2vec-lstm      & {\cellcolor[rgb]{0.655,0.855,0.714}}10.39\% & {\cellcolor[rgb]{0.973,0.459,0.467}}-15.63\% & {\cellcolor[rgb]{0.957,0.976,0.973}}1.47\%  & {\cellcolor[rgb]{0.973,0.463,0.471}}-15.50\% & {\cellcolor[rgb]{0.902,0.953,0.925}}3.11\%   \\
                        & word2vec-sum       & {\cellcolor[rgb]{0.882,0.945,0.91}}3.67\%   & {\cellcolor[rgb]{0.973,0.412,0.42}}-17.09\%  & {\cellcolor[rgb]{0.984,0.969,0.98}}0.00\%   & {\cellcolor[rgb]{0.973,0.42,0.427}}-16.75\%  & {\cellcolor[rgb]{0.984,0.843,0.855}}-3.84\% \\ 
                        \hline
\end{tabular}
}
\caption{Difference in training and validation accuracy across different levels of mutation}
\label{tab:generalisationGap}
\end{table*}




\end{document}